%% file: CNNJournal.tex
\pdfoutput=1

\documentclass[letterpaper, 10 pt, conference]{IEEEtran}  % Comment this line out if you need a4paper

\IEEEoverridecommandlockouts                              % This command is only needed if 
\usepackage{epsfig} % for postscript graphics files
\usepackage{mathptmx} % assumes new font selection scheme installed
\usepackage{times} % assumes new font selection scheme installed
\usepackage{amsmath} % assumes amsmath package installed
\usepackage{amssymb}  % assumes amsmath package installed
\usepackage{graphicx} 
\usepackage{graphics} 
\usepackage{grffile} 
\usepackage[T1]{fontenc}
\usepackage{amsfonts}
\usepackage{booktabs}
\usepackage{siunitx}
\usepackage{caption} 
\usepackage{booktabs}
\usepackage{graphicx}
\usepackage{algorithmic}
\usepackage[ruled]{algorithm2e}
\usepackage{float}

\usepackage{mwe,tikz}\usepackage[percent]{overpic}

\usepackage{subcaption}
\usepackage{array,   tabularx, makecell, booktabs}%
\usepackage{geometry}
\usepackage{comment}

\DeclareMathOperator*{\argmax}{arg\,max}
\DeclareMathOperator*{\argmin}{arg\,min}
\title{\LARGE \bf
Arena-Rosnav: Towards Deployment of Deep-Reinforcement-Learning-Based Obstacle Avoidance into Conventional Autonomous Navigation Systems
}

\author{Linh K{\"a}stner$^{1}$\thanks{$^{1}$Linh K{\"a}stner, Teham Buiyan, Lei Jiao, Tuan Anh Le, Xinlin Zhao, Zhengcheng Shen and Jens Lambrecht are with the Chair Industry Grade Networks and Clouds, Faculty of Electrical Engineering, and Computer Science,				
		Berlin Institute of Technology, Berlin, Germany
		{\tt\small linhdoan@tu-berlin.de}}, Teham Buiyan$^{1}$, Lei Jiao$^{1}$, Tuan Anh Le$^{1}$, Xinlin Zhao$^{1}$,\\ Zhengcheng Shen$^{1}$ and Jens Lambrecht$^{1}$
}

\begin{document}

\maketitle
\thispagestyle{empty}
\pagestyle{empty}

\input{ieeeconf/content/0-abstract}
\input{ieeeconf/content/1-introduction}

% \newgeometry{top=0.75in,bottom=0.75in,right=0.75in,left=0.75in}

\input{ieeeconf/content/2-Related_Works}

\input{ieeeconf/content/3-methodology}

\input{ieeeconf/content/4-evaluations}

\input{ieeeconf/content/5-conclusion}

%\section*{Acknowledgement}
%We acknowledge help with the production of the spot welds and with the chisel test by Hubert Suwala.

\addtolength{\textheight}{-1cm} 

%%%%%%%%%%%%%%%%%%%%%%%%%%%%%%%%%%%%%%%%%%%%%%%%%%%%%%%%%%%%%%%%%%%%%%%%%%%%%%%%

%%%%%%%%%%%%%%%%%%%%%%%%%%%%%%%%%%%%%%%%%%%%%%%%%%%%%%%%%%%%%%%%%%%%%%%%%%%%%%%%

%%%%%%%%%%%%%%%%%%%%%%%%%%%%%%%%%%%%%%%%%%%%%%%%%%%%%%%%%%%%%%%%%%%%%%%%%%%%%%%%

%%%%%%%%%%%%%%%%%%%%%%%%%%%%%%%%%%%%%%%%%%%%%%%%%%%%%%%%%%%%%%%%%%%%%%%%%%%%%%%%
\typeout{}
\bibliographystyle{IEEEtran}
\bibliography{CNNJournal}

\end{document}

%% file: ieeeconf/content/0-abstract.tex
% !TeX encoding = utf-8
% !TeX language = en_GB
% !TeX spellcheck = en_GB
% !TeX root = paper.tex

\begin{abstract}
Recently, mobile robots have become important tools in various industries, especially in logistics. Deep reinforcement learning emerged as an alternative planning method to replace overly conservative approaches and promises more efficient and flexible navigation. However, deep reinforcement learning approaches are not suitable for long-range navigation due to their proneness to local minima and lack of long term memory, which hinders its widespread integration into industrial applications of mobile robotics.
In this paper, we propose a navigation system incorporating deep-reinforcement-learning-based local planners into conventional navigation stacks for long-range navigation. Therefore, a framework for training and testing the deep reinforcement learning algorithms along with classic approaches is presented. We evaluated our deep-reinforcement-learning-enhanced navigation system against various conventional planners and found that our system outperforms them in terms of safety, efficiency and robustness.

\end{abstract}

%% file: ieeeconf/content/1-introduction.tex
% !TeX encoding = utf-8
% !TeX language = en_GB
% !TeX spellcheck = en_GB
% !TeX root = paper.tex

%%%%%%%%%%%%%%%%%%%%%%%%%%%%%%%%%%%%%%%%%%%%%%%%%%%%%%%%%%%%%%%%%%%%%%%%%%%%%%%%
\section{Introduction}
Mobile robots have gained significant importance due to their flexibility and the variety of use cases they can operate in \cite{alatise2020review}. At the same time, the environments in which mobile robots operate have become increasingly complex, with multiple static and dynamic obstacles like humans, fork lifts or robots. Reliable and safe navigation in these highly dynamic environments is essential in the operation of mobile robotics \cite{robla2017working}. Whereas classic planning approaches can cope well with static environments, reliable obstacle avoidance in dynamic environments remains a big challenge.
Typically, current industrial approaches employ hand-engineered safety restrictions and navigation measures \cite{sisbot2005navigation}, \cite{qian2010socially}. In particular, in environments that employ a variety of different actors, industrial robots are often programmed to navigate overly conservatively, or to avoid such areas completely due to stringent restrictions.
%which would have to be raised. 
\begin{figure}[!h]
	\centering
	\includegraphics[width=3in, height=2.3in]{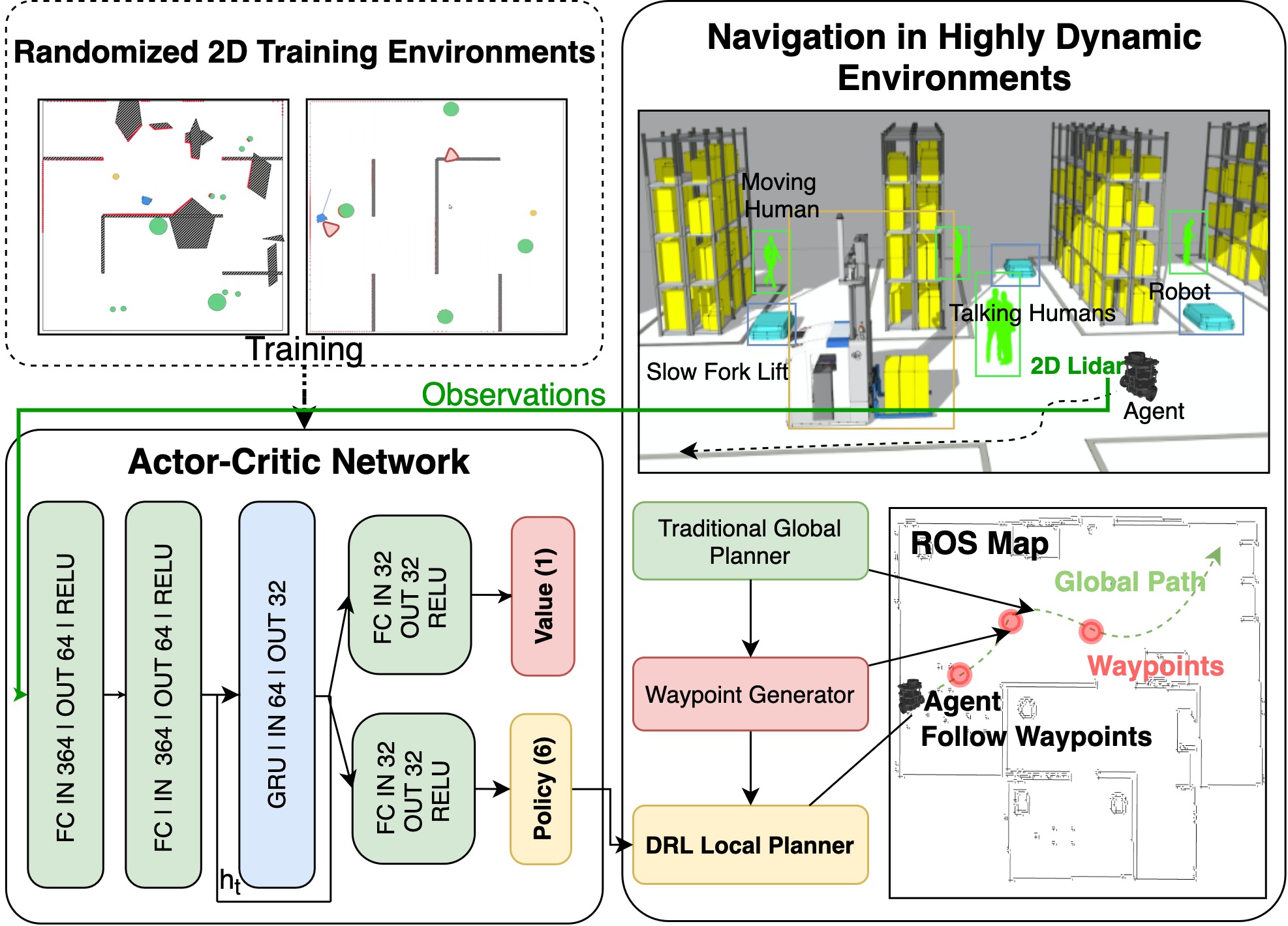}
	\caption{This work provides a platform to train and test learning-based obstacle avoidance approaches along with conventional global and local planners. This way, DRL-based obstacle avoidance approaches are made comparable against conventional approaches and feasible in industrial navigation systems}
	\label{intro}
\end{figure}
However, hand designing the navigation behavior in dense environments is tedious and not always intuitive, due to not only the unpredictability of humans, but also volatile behavior patterns of other actors.\\
Deep Reinforcement Learning (DRL) emerged as an end-to-end learning approach, which directly maps raw sensor outputs to robot actions and has shown promising results in teaching complex behavior rules, increasing robustness to noise and generalizing new problem instances. Thus, a variety of literature incorporates DRL into robot navigation \cite{chiang2019learning}-\cite{chen2017socially}. However, a main bottleneck is its limitation for local navigation, due to a lack a long term memory and its myopic nature \cite{dugas2020navrep}. Efforts to integrate recurrent networks to mitigate this issue result in tedious training and limited payoff. 
In this paper, we address current limitations of integration of DRL into industrial robotic navigation systems for long-range navigation in highly dynamic environments. We propose a navigation framework, which is fully integrated into the robot operating system (ROS) and enables the deployment of DRL-based planning as part of existing navigation stacks. As an interconnection entity, an intermediate way-point planner is provided, which connects DRL with conventional global planning algorithms like RRT, $A^*$ or Dikstra.
Furthermore, the platform serves as a both training and testing playground and provides a 2D simulator in which training a DRL-based local planner using robotic interfaces is made feasible. The main contributions of this work are the following:
\begin{itemize}
    \item Proposal of a framework to train and integrate learning-based approaches into conventional robot navigation stacks. 
    \item Proposal of an intermediate way-point generator for interconnection between DRL and conventional global planning modules.
    \item Integration of state-of-the-art obstacle avoidance approaches into the robot operating system (ROS) for extensive evaluation of navigation systems.
\end{itemize}
The paper is structured as follows. Sec. II begins with related works followed by the methodology in Sec. III. Subsequently, the results and evaluations are presented in Sec IV. Finally, Sec. V provides a conclusion and outlook.
We made our code open-source at https://github.com/ignc-research/arena-rosnav.

%% file: ieeeconf/content/2-Related_Works.tex
\section{Related Works}
Autonomous navigation of mobile robots has been extensively studied in various research publications. While current robot navigation planners work well in static environments or with known obstacle positions, highly dynamic environments still pose an open challenge.
Current mobile navigation approaches often employ hand-engineered measures and rules \cite{lam2010human}, \cite{guzzi2013human}, safety thresholds \cite{robla2017working}, dynamic zones contemplating to social conventions \cite{truong2016dynamic} or the complete avoidance of areas known to be problematic \cite{qian2010socially}. These approaches are reaching their limits when the exact models are unknown or become complex for convoluted scenarios, which in term could lead to longer paths, waiting times or complete failure. Furthermore, in highly dynamic environments employing numerous behavior classes, these hand-engineered rules could become tedious and computationally intensive. DRL has emerged as an end-to-end approach with potential to learn navigation in dynamic environments. Thus, various publications utilize DRL for path planning and obstacle avoidance \cite{chiang2019learning}, \cite{faust2018prm}, \cite{francis2020long}, \cite{shi2019end}.
Shi et al. \cite{shi2019end} proposed a DRL-based local planner utilizing the Asynchronous Advantage Actor Critic (A3C) approach and achieved end-to-end navigation behavior purely from sensor observations. The researchers transfer the approach towards a real robot and demonstrate its feasibility. 
Works from Sun et al. \cite{sun2020inverse}, Truong et al. \cite{truong2017toward} and Ferrer et al. \cite{ferrer2019anticipative} incorporate a human motion prediction algorithm into the navigation to enhance environmental understanding and adjust the navigation for crowded environments. A socially aware DRL-based planner (CADRL) was proposed by Chen et al. \cite{chen2017socially}. The researchers introduce reward functions and aim to teach the robot social norms like keeping to the right side of the corridor. Everett et al. \cite{everett2018motion} extend CADRL to include the handling of a dynamic amount of persons using an Long-Term-Short-Memory module and modify the approach to solve human randomness. 
Chen et al. \cite{chen2019crowd} propose an obstacle avoidance approach using DRL by modeling human behavior. 
Despite the success of DRL for local navigation and obstacle avoidance, integration of these algorithms into conventional robotic navigation systems is still an open frontier, since most obstacle avoidance (OA) algorithms are purely developed as a numerical optimization problem. On that account, our proposed framework aims to provide a platform to employ learning-based OA approaches directly for robot navigation.
Similar to our work, Dugas et al. \cite{dugas2020navrep} propose a platform consisting of a 2D simulation environment to train and test different DRL algorithms. The researchers integrate a variety of different DRL-based navigation approaches into their platform and compare them against one another. The work focused on the general navigation aspect of DRL-based planners and only tested on one scenario with dynamic obstacles. Our work extends this by incorporating the global planner for long range navigation and specifically deploy the planners on multiple challenging, highly dynamic environments. 
Most similar to our work, Gundelring et al. \cite{guldenringlearning} first combine a DRL-based local planner with a conventional global planner and demonstrate promising results. A limitation of that work is that the researchers only employ simple sub-sampling of the global path and no replanning capability which leads to hindrance when there are multiple humans blocking the way. In this work, we follow a similar approach but introduce an intermediate planner to spawn way-points more intelligent and add replanning functionality based on spatial and time horizons. 
\begin{comment}
Thus, our work will provide a platform to integrate both DRL-based local planning into the navigation stack to make DRL feasible for industrial application.

\end{comment}

%% file: ieeeconf/content/3-methodology.tex
\section{Methodology}
In this chapter we will present the methodology of our proposed framework. Since DRL-based methods are myopic and not suitable for long range navigation tasks \cite{dugas2020navrep}, our work proposes a combination with traditional global planners to make DRL feasible for industrial applications. The system design is described in Fig. \ref{intro}. Part of our contribution is an interconnection way-point generator which connects the two planners. The platform serves as both a training and testing environment and is integrated into the ubiquitously used ROS. In the following, each sub-module is described in detail. 
\subsection{Intermediate Planner - Waypoint Calculation}
In order to accomplish long range navigation while utilizing a DRL-based local planner, a hierarchical path planning is implemented consisting of an intermediate planner as an interconnection between the traditional global planner and the DRL-based local planner. $A^*$ search is used as a global planner to find a near optimal global path. Instead of a simple sub-sampling as proposed in Gundelring et al. \cite{guldenringlearning}, which is not flexible for stuck situations, we introduce a spatial horizon dependent on the robots position. The algorithm is described in Algorithm \ref{subsampling}.
\begin{figure}[!h]
\centering
\includegraphics[width=0.2\textwidth]{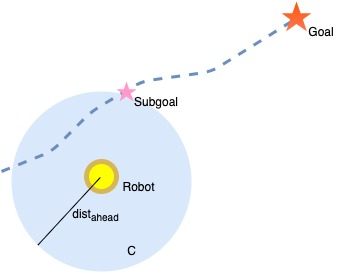}
\caption{Subgoal selection. Given a global path, our intermediate planner, will calculate a subgoal based on the spatial horizon $dist_{ahead}$.}
\label{fig:subgoal}
\end{figure}

\begin{algorithm}[!h]
  \SetAlgoLined
  \KwIn{ Global path $\pi_{g}$, Robot position $p_r$}
  \KwOut{Subgoal $g_{sub}$ }
  Parameter: $d_{ahead}$\;
  Find the set of intersection points of $R(p_r,d_{ahead})$ and global path $\pi_{g}$ as $\Phi$\;
  \eIf{$\Phi$ is not empty}
  {select the intersection point near to the global goal as $g_{sub}$}
  {
    call GLOBAL REPLAN $\rightarrow$ new global path $\pi_{g}^{new}$\;
    Subgoal calculation$(\pi_{g}^{new},p_r)$\;
  }
  \caption{Subgoal calculation}
  \label{subsampling}
\end{algorithm}

Let the global path $\pi_g$ be an array Y consisting of $N_t$ poses from which we sample a subset $Y^{\text{subgoals}_i }\subset X^{poses}= \{x_i, \dots, x_N\}$ of local goals based on the robot's position $p_r$ and a look-ahead distance $d_{ahead}$. Provided the global path and the current robot's position, a sub-goal within the range of look-ahead distance is selected as the local goal for the DRL local planner. Thus, the subgoal is dynamic and moves with the robot's position. This makes the approach more flexible in situations with multiple unknown obstacles. Fig. \ref{fig:subgoal} illustrates our approach.
The subset of goals can be observed as local sub-goals $x_i = p_g$ and be given as input into the DRL agent. Additionally, we integrate a replanning mechanism to rework the global path when the robot is off-course for a certain distance from the global path or if a time limit $t_{lim}$ is reached without movement, e.g. when the robot is stuck. In that case, a global replanning will be triggered and a new sub-goal will be calculated based on the new global plan and current robot position.

\subsection{Deep-Reinforcement-Learning-based Local Navigation}
We formulate the robot navigation problem as a partially observable Markov decision problem (POMDP) that is formally describable by a 6-tuple $(S,A,P,O,R, \gamma)$
where S are the robot states, A the action space, P the transition probability from one state to another, R the reward, O the observations and $\gamma$ the discount factor which we keep at 0.9. Next, we formulate following task constraints for a collision free navigation towards the goal:
\begin{align}
   C_{j,i,t}(s,a) = 
   \begin{cases} 
        |p_r - p_g|_2 < d_{goal} &  \textit{goal constraint}   \\
        \max(O_{s,t}) > d_r  & \mbox{ \textit{collision constraint} } \\
        \argmin [t | s, \pi]  & \mbox{ \textit{time constraint} } \\
    \end{cases}  
\end{align}
Where $d_{goal}$ is the goals radius, $p_r$ is the robot's position, $p_g$ is the goal's position , O is the robot's observation and $d_r$ is the robot's radius.
To solve the POMDP, we employ policy-based model-free actor critic reinforcement learning to learn a policy which maximizes a cumulative reward dependent on the state action pair. It does so by calculating the policy gradients, which optimize both a value (critic) and policy function (actor).
The value function is calculated by accumulating the expected rewards at each time-step given an action performed while following policy $\pi$
\begin{align}
    V^*(s_0) = \argmax \mathbb{E} \left[ \sum_{t=0}^T \gamma R(s_t,\pi^*(s_t)) \right]
\end{align}
Subsequently, the Bellman equation can be iterated over the value to calculate the optimal policy function $\pi(s_{t+1})$:
\begin{align}
    \pi^* (s_{t+1}) = \argmax & R(s_t, a) + \\
    &\int_{s_{t+1}} P(s_t,s_{t+1} | a ) \cdot V^*(s_{t+1}) \,ds_{t+1} 
\end{align}

\subsubsection{Training Algorithm}
To train our DRL agent, we utilize an A3C policy gradient method whose output is split into a policy (actor) and a value (critic) function. Contrary to most Deep-Q-Learning algorithms, the approach is an on-policy training method in which the agent has to be trained on the current status of the network. To avoid sample inefficiency, actor critic approaches employ asynchronous training of multiple actors simultaneously, each with its own environment $N$. The knowledge gained from all actors is then concatenated into the final output. For an agent with a state $s_t$ and an action $a_t$ at a time step $t$, 
the policy output $\mu (s_i,t)$, is a probability distribution of all actions, whereas the advantage output $V_\theta (s_i)$ assesses the advantage of executing the actions proposed by the policy. For our case, we forward the observations of $N$ agents simultaneously as depicted in Fig. \ref{nn}. The current observations are passed into the actor-critic network, which returns the state value $V_\theta (s_i)$ and the policy $\pi_\theta$ upon which the simulation step is executed. The obtained reward $r_i$ is saved into the buffer $R$. These values are used to calculate the value loss $L_{value}$ and the policy loss $L_{policy}$ with a mean squared error. Softmax is applied to obtain the policy values into probabilities of each action. Finally, the policy-gradient ($\mu (s_i,t)$) and value-gradient ($V_\theta (s_i)$) are calculated using the respective losses. The network parameters are updated accordingly. We use an Adam optimizer with a learning rate of 0.0001 and an $\epsilon$ of 0.003. The discount factor is set to 0.9.
The training algorithm is described in Algorithm \ref{alg222}.

\begin{algorithm}[!h]
  \SetAlgoLined
  $L \leftarrow \mbox{new Array}(B_{\mbox{size}})$\;
  $\pi_\theta , V_\theta (s_i) \leftarrow \mbox{net(observations)}$\;
  \For{$i=t-1 ... t_{start}$}{
    \For{$N$} {
      \eIf{\mbox{episode ended}}{
      $R = 0$
      }{$R = V_\theta (s_i) $}
      $\mbox{R} \leftarrow \mbox{rewards} + \gamma R$
      $L_{value} \leftarrow (R-V_\theta(s_i))^2$
      $L_{policy} \leftarrow \mbox{mse}(\mbox{softmax}(\pi_\theta(s_i)) (R-V_\theta(s_i)))$
      $\frac{\partial}{\partial v} \leftarrow \frac{\partial}{\partial v}  + \frac{\partial L_{value}(\theta_v)}{\partial \theta_v} $
      $\frac{\partial}{\partial\theta_\pi} \leftarrow \frac{\partial}{\partial\theta_\pi}  + \nabla_\theta log \pi_\theta(a_i|s_i) L_{policy} $
    }
  }
  
  \caption{Training Algorithm of A3C}
  \label{alg222}
\end{algorithm}

\subsubsection{Reward System}
We formulate our reward system as following. First, we define $p_{g}$ as the goal position and $p_r$ being the agent's position.
The total reward $R(s_t,a)$ is calculated as the sum of all sub-rewards.
\begin{align}
    R(s_t,a) = [r_{s}^t + r_{c}^t + r_{d}^t + r_{p}^t + r_{m}^t]
\end{align}
\begin{align*}
r_{s}^t&=\begin{cases}15&|p_r - p_g|_2 < d_{g}\\0&otherwise\end{cases} \qquad
r_{m}^t =\begin{cases}0& otherwise\\-0.01& \Delta r =0 \end{cases}  \\
r_{d}^t& =\begin{cases}0&otherwise\\-0.15& d_r < d_{safe}\end{cases} \qquad
r_{c}^t=\begin{cases}0& otherwise\\-10& \max(O_{s,t}) < d_r\end{cases}
\end{align*}
\begin{align*}
r_{p}^t& =\begin{cases}w_{p}*d^t& d^t>=0,\;d^t=d_{ag}^{t-1}-d_{ag}^{t}\\w_{n}*d^t& else\end{cases}
\end{align*}
Here, $r_s$ is the success reward, $r_d$ is the distance reward, $r_p$ is the progress reward, $r_m$ is the move reward, $r_d$ is the danger reward to keep the agent from a safety distance $d_{safe}$ and $r_c$ is the collision reward. We set $w_{p} = 0.25, w_{n} = 0.4 $. If the robot moves away from the dynamic obstacle, it receives an additional reward. 

\subsubsection{Neural Network Architecture and Agent Design}
\begin{figure}[]
    \centering
	\includegraphics[width=0.33 \textwidth]{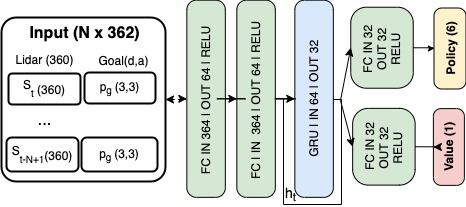}
	\caption{ Neural network architecture. We utilize an on policy, actor critic agent with a gated recurrent unit. As input we forward the down-sampled LIDAR scan observations into the network. For both, actor and critic network, we use the same head.}
	\label{nn}
\end{figure}

Our neural network architecture is illustrated in Fig. \ref{nn}. It consists of two networks, one for the value and one for the policy function. The input to our network is the 360 degree LIDAR scan, which we down-sampled to 344 values as well as the goal position and angle with respect to the robot. After two fully connected layers, we add a gated recurrent unit (GRU) as memory module to also consider past observations that are stored in a buffer. Subsequently, we train on a continuous action state for more flexibility and smoothness of actions \cite{faust2018prm}. We formalize the action space $A$ as following:
\begin{align}
    a = \{v_{lin}& , v_{ang} \} \\
    v_{lin} \in [0, 0.3] m/s, \quad
    & v_{ang} \in [-2.7, 2.7] m/s
\end{align}

\subsubsection{Training Setup}
The agent is trained on randomized environments which are depicted in Fig. \ref{intro}. Walls and static obstacles are spawned randomly after each episode. In the same manner, dynamic obstacles are spawned at random positions and move randomly. This way, we mitigate over-fitting issues and enhance generalization. Application interfaces to include more complex obstacle models and shapes are provided. Curriculum training is adapted which spawns more obstacles once a success threshold is reached and less obstacles if the agents success rate is low. the threshold is defined as the sum over the mean reward. The training was optimized with GPU usage and trained on a NVIDIA RTX 3090 GPU. Training time took between 8h and 20h to converge. The hyperparamters are listed in Table \ref{tablehyper} in the Appendix.

\subsubsection{Integration of Obstacle Avoidance and Alternative Approaches}
Our framework provides interfaces to integrate obstacle avoidance methods into our navigation system. For comparison of our DRL-based local planner, we integrate a variety of optimization-based obstacle avoidance algorithms into ROS. This way, even purely numerical-optimization-based obstacle avoidance algorithms are made feasible for deployment on robotic navigation systems. Particularly, we included Timed Elastic Bands (TEB), Dynamic Windows Approach (DWA), Model Predictive Control (MPC), as well as learning-based obstacle avoidance methods CADRL and CrowdNav. In the next chapter we will use these approaches as baseline for our DRL-based local planner. 

\begin{comment}

The baselines are listed in Table \ref{baselines}.
\begin{table}[!h]

	\centering
	\setlength{\tabcolsep}{1pt}
	\renewcommand{\arraystretch}{1.3}
	\caption{Baseline Local Planners}
	\begin{tabular}{p{1.4cm}|p{2.40cm}p{1.2cm}p{3.2cm}}
		\hline
		\hline
		Agent    & Input & Output & Comments  \\ \hline
		DWA & Lidar, costmap & twist & directly deployable \\
		TEB & Lidar, costmap & twist &directly deployable \\
		MPC & Lidar costmap & twist &directly deployable \\ 
		CADRL & obstacle position & angle &does not consider static obstacles   \\
		Arena2d & Lidar & twist &working only on lidar \\
	    \hline
		\hline
		
	\end{tabular}

	\label{baselines}
	
\end{table}

\end{comment}

%% file: ieeeconf/content/4-evaluations.tex
\begin{figure*}[!h]
	
	\begin{subfigure}{0.29\textwidth}(a)
		\includegraphics[width=\linewidth]{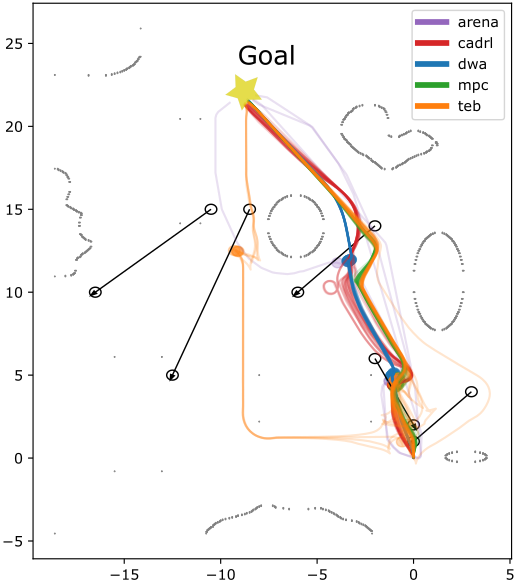}
		
		\label{fig:1}
	\end{subfigure}\hfil % <-- added
	\begin{subfigure}{0.29\textwidth }(b)
		\includegraphics[width=\linewidth]{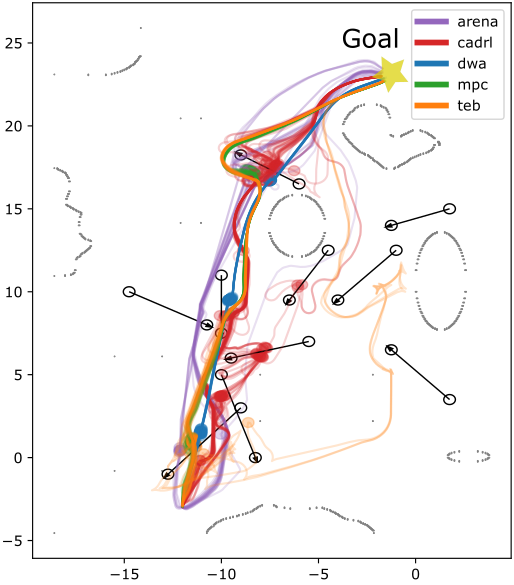}
		
		\label{fig:2}
	\end{subfigure}\hfil % <-- added
	\begin{subfigure}{0.29\textwidth}(c)
		\includegraphics[width=\linewidth]{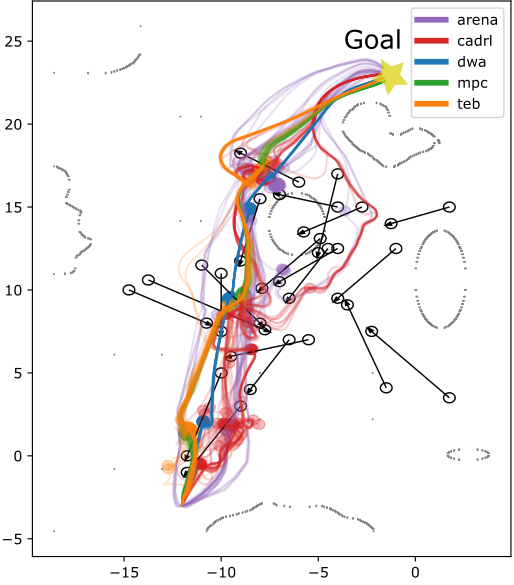}
		
		\label{fig:3}
	\end{subfigure}
	
	\medskip
	\begin{subfigure}{0.30\textwidth}(d)
		\includegraphics[width=\linewidth]{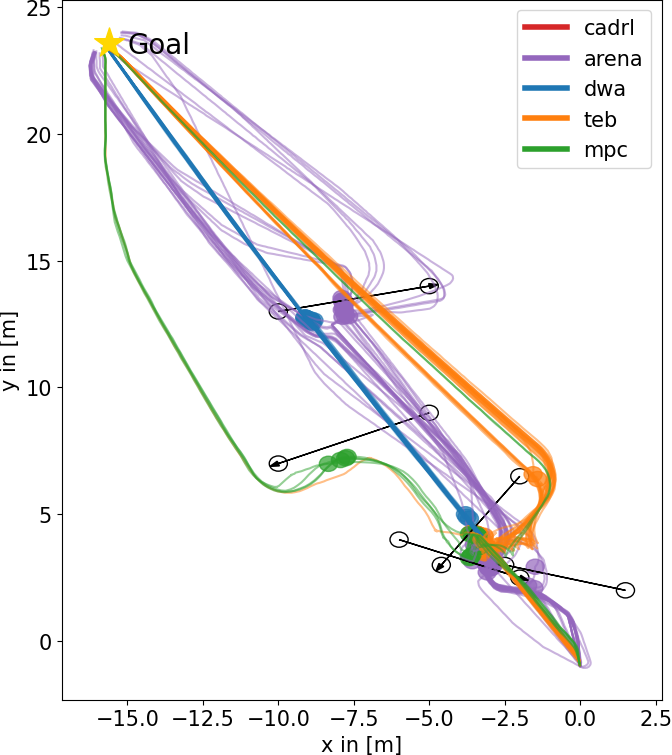}
		
		\label{fig:4}
	\end{subfigure}\hfil % <-- added
	\begin{subfigure}{0.29\textwidth}(e)
		\includegraphics[width=\linewidth]{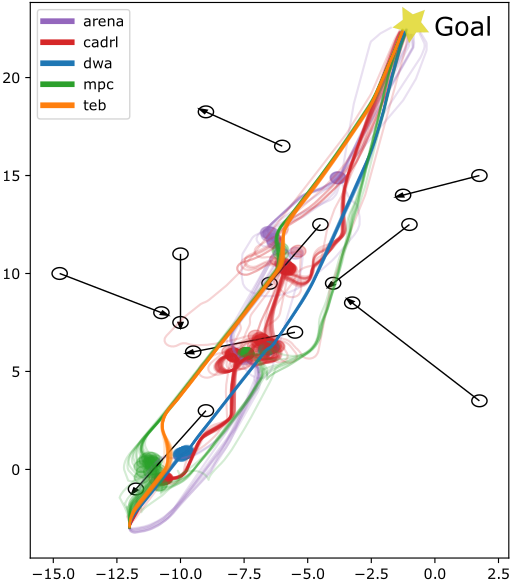}
		
		\label{fig:5}
	\end{subfigure}\hfil % <-- added
	\begin{subfigure}{0.29\textwidth}(f)
		\includegraphics[width=\linewidth]{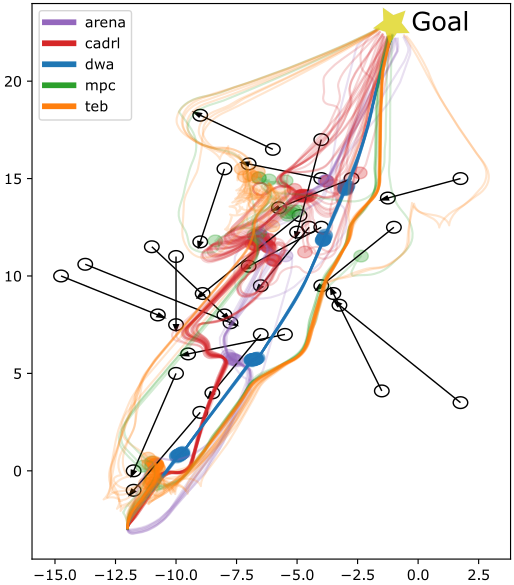}
		
		\label{qual}
	\end{subfigure}
	\caption{Trajectories of all planners on different test scenarios with obstacle velocity $v_{obs}=0.3 m/s$. Upper row: office map, lower row: plain test area. (a) 5 dynamic obstacles approaching robot from side, (b) 10 dynamic obstacles intersecting horizontally and blocking corridor, (c) 20 dynamic obstacles intersecting horizontally, (d) 5 obstacles randomly intersecting the robots path on an empty map, (e) 10 obstacles intersecting horizontally, (f) 20 obstacles intersecting horizontally.}
	\label{quali}
\end{figure*}

\section{Results and Evaluation}
In the following chapter, we will present our experiments, the results and evaluations. This includes an extensive comparison with 3 state of the art model-based local planners TEB \cite{rosmann2015timed}, DWA \cite{kurtz2019toward} and MPC \cite{rosmann2019time} as well as a recent obstacle avoidance approache based on DRL: CADRL \cite{everett2018motion}, with our proposed DRL local planner which we denote as ARENA. 

\subsection{Qualitative evaluation}
We conducted experiments on two different maps in which we created a total of seven different test scenarios of increasing difficulty. In each scenario, the obstacle velocities are set to 0.1 m/s. More extensive evaluations on the impact of obstacle velocities are presented in the quantitative evaluations. As a global planner, $A^*$ is used for all approaches. For the integrated OA approaches and the DRL-based planners, the global path is sampled by the intermediate way-point generator using the time and location horizon of $t_{lim}=4s$ and $d_{ahead}=1.55m$. For the model-based approaches, we utilize the ROS navigation stack interface. Localization is assumed to be perfectly known. For each planner, we conduct 100 test runs on each scenario. The qualitative trajectories of all planners on each scenario are illustrated in Fig. \ref{quali} (a)-(f). The intensity of the trajectories indicates their frequency.

\begin{comment}
\begin{figure*}[!h]
\centering
\includegraphics[width=6.9in]{ieeeconf/content/img/histnew.jpg}
\caption{Upper row: plain test area, lower row: office map. (a) dynamic obstacles approaching robot from the front, (b) 5 dynamic obstacles intersecting horizontally, (c) dynamic obstacles approaching in a zig-zag pattern from front, (d) 10 obstacles randomly intersecting the robots path, (e) 5 obstacles on office map moving randomly, (f) 10 obstacles moving randomly and blocking corridors, (g) 20 obstacles moving randomly. Collisions are visualized as circles with the respective colors. The intensity of the circles indicates how frequently collisions occurred. The obstacle movements are represented as vectors from start to end position.}
\label{quali}
\end{figure*}
\end{comment}

\subsubsection{Robot Trajectories}
For scenarios that include more than five obstacles, DWA and TEB struggle due to the slow reaction time of the local planners. Particularly, the trajectories of the TEB planner (yellow) show stops and backwards movements once the TEB planner perceives an obstacle or at direction changes and moves backwards which is inefficient and sometimes too slow, especially if there are multiple obstacles approaching from different directions. Nevertheless, when the obstacles are not directly confronting the agent, these approaches can attain robust and efficient path planning which follows the global planner consistently as can be observed by the trajectories being consistent for most test runs. Notable is the high performance of MPC in the office map where it accomplishes the best results in terms of obstacle avoidance and path robustness out of the model-based approaches. The trajectories of the MPC planners are consistent and do not include as many outliers. However, in the empty map, more outliers are observed in situations where the obstacle is directly approaching the agent (Fig. 4(d)), whereas the TEB planner performs slightly better with less outliers. Our Arena planner as well as CADRL manages to keep a direct path towards the goal and still accomplishes less collisions compared to the model-based planners. The collisions are visualized as circles with varying intensity based on how frequently collisions occurred. CADRL avoids dynamic obstacles in a larger circular movement compared with our ARENA planner which leads to longer trajectories (red). It is observed that our planner reacts to approaching obstacles already from a far distance while trying to follow the global path once the obstacle is avoided to ensure efficiency. Contrarily, conventional planners react slowly to approaching obstacles and rely a more frequent replanning of the global path. Particularly, the TEB planner employs an inefficient stop and backward motion and a large amount of inefficient roundabout paths (Fig. 4 (a),(b),(f)). Our proposed ARENA planner accomplishes a robust path to the goal while accomplishing a reliable collision avoidance. Additionally, our intermediate planner replans more efficiently in highly dynamic environments by considering the introduced spatial and time horizons which triggers a replanning only when the robot becomes stuck for more than four seconds. In general, it is concluded that the model-based approaches follow the global path more consistently while learning-based approaches make decisions earlier thus more often got off-track when avoiding the obstacles.
\begin{figure*}[!h]
	\centering % <-- added
	\begin{subfigure}{0.3\textwidth}(a)
		\includegraphics[width=\linewidth]{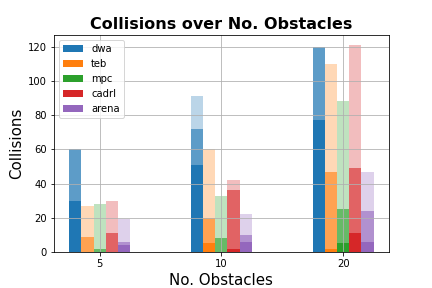}
		
		\label{fig:1}
	\end{subfigure}\hfil % <-- added
	\begin{subfigure}{0.3\textwidth}(b)
		\includegraphics[width=\linewidth]{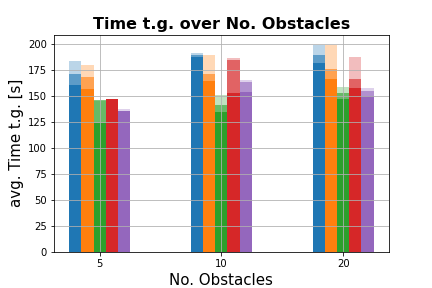}
		
		\label{fig:2}
	\end{subfigure}\hfil % <-- added
	\begin{subfigure}{0.3\textwidth}(c)
		\includegraphics[width=\linewidth]{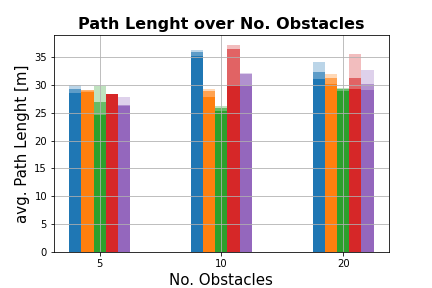}
		
		\label{fig:3}
	\end{subfigure}
	
	\medskip
	\begin{subfigure}{0.3\textwidth}(d)
		\includegraphics[width=\linewidth]{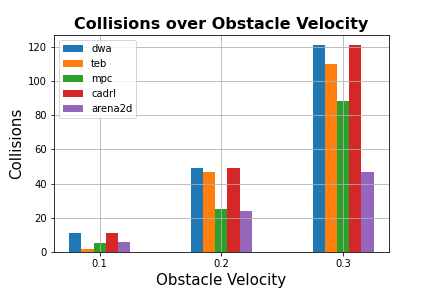}
		
		\label{fig:4}
	\end{subfigure}\hfil % <-- added
	\begin{subfigure}{0.3\textwidth}(e)
		\includegraphics[width=\linewidth]{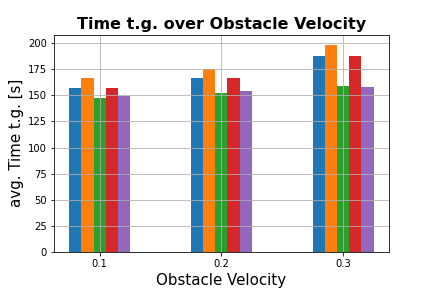}
		
		\label{fig:5}
	\end{subfigure}\hfil % <-- added
	\begin{subfigure}{0.3\textwidth}(f)
		\includegraphics[width=\linewidth]{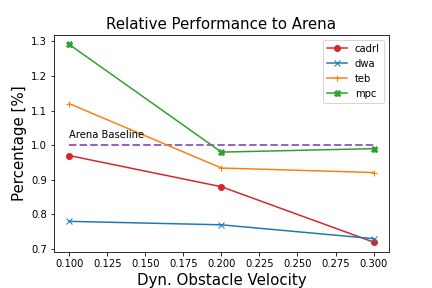}
		
		\label{fig:6}
	\end{subfigure}
	\caption{Results from test-runs on the office map. (a) Collisions over number of obstacles and $v_{obs}$, (b) average time to reach goal over number of obstacles and $v_{obs}$, (c) Average path length over number of obstacles and $v_{obs}$. With $v_{obs}=(0.1,0.2,0.3) m/s$, while a high opacity indicates a lower velocity.(d) Collisions over obstacle velocities, (e) average time to reach goal over obstacle velocities (lower is better), (f) Relative performance to ARENA planner.}
	\label{quantgraph1}
\end{figure*}

\begin{figure*}[!h]
	\centering % <-- added
	\begin{subfigure}{0.32\textwidth}(a)
		\includegraphics[width=\linewidth]{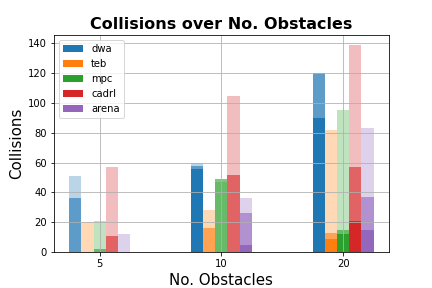}
		
		\label{fig:1}
	\end{subfigure}\hfil % <-- added
	\begin{subfigure}{0.32\textwidth}(b)
		\includegraphics[width=\linewidth]{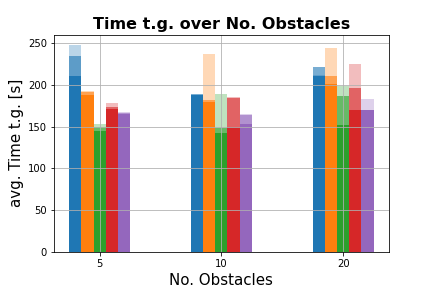}
		
		\label{fig:2}
	\end{subfigure}\hfil % <-- added
	\begin{subfigure}{0.32\textwidth}(c)
		\includegraphics[width=\linewidth]{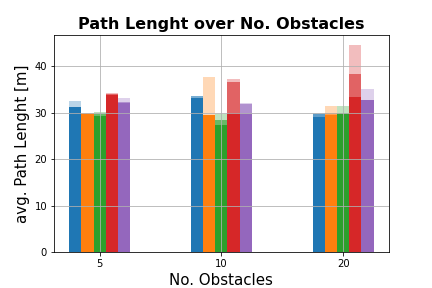}
		
		\label{fig:3}
	\end{subfigure}
	
	\medskip
	\begin{subfigure}{0.32\textwidth}(d)
		\includegraphics[width=\linewidth]{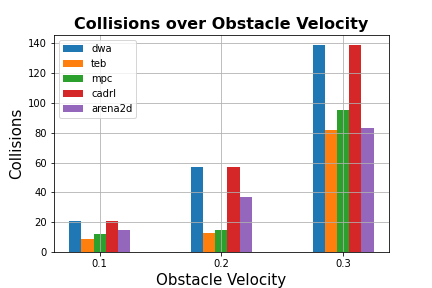}
		
		\label{fig:4}
	\end{subfigure}\hfil % <-- added
	\begin{subfigure}{0.32\textwidth}(e)
		\includegraphics[width=\linewidth]{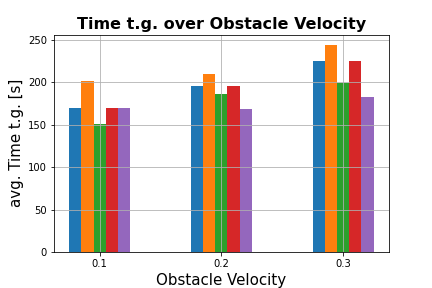}
		
		\label{fig:5}
	\end{subfigure}\hfil % <-- added
	\begin{subfigure}{0.32\textwidth}(f)
		\includegraphics[width=\linewidth]{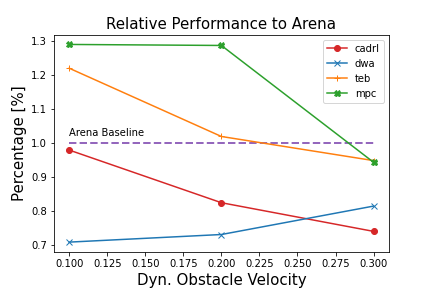}
		
		\label{fig:6}
	\end{subfigure}
	\caption{Results from test-runs on the empty map. (a) Collisions over number of obstacles and $v_{obs}$, (b) Average time to reach goal over number of obstacles and $v_{obs}$, (c) Average path length over number of obstacles and $v_{obs}$. With $v_{obs}=(0.1,0.2,0.3) m/s$, while a high opacity indicates a lower velocity. (d) Collisions over obstacle velocities, (e) Average time to reach goal over obstacle velocities (lower is better), (f) Relative performance to ARENA planner.}
	\label{quantgraph2}
\end{figure*}

\begin{table*}[!h]
	\centering
	\setlength{\tabcolsep}{3pt}
	\renewcommand{\arraystretch}{1}
	\caption{Quantitative evaluations}
	\begin{tabular}{p{2.2cm}|p{1cm}p{1cm}p{1cm}p{1cm}|p{1cm}p{1cm}p{1cm}p{1cm}|p{1cm}p{1cm}p{1cm}p{1cm}}
		\hline
		\hline
		& Time [s]& Path [m]  & Collisions &  Success & Time [s]& Path  [m]  & Collisions &  Success & Time [s]& Path [m]  & Collisions &  Success \\ \hline
		
		\textbf{$v_{obs}$ 0.1m/s}& \multicolumn{4}{c}{\bfseries 5 dyn. Obstacles} & \multicolumn{4}{c}{\bfseries 10 dyn. Obstacles} & \multicolumn{4}{c}{\bfseries 20 dyn. Obstacles} \\ 
		\cline{1-13}
		CADRL & 158.69 & 31.07  & 0     & 100   & 151.21 & 29.97 & \textbf{1}    & 100  & 163.5   & 31.30 & 16   & 95   \\
		\textbf{ARENA (Ours)}& 150.24  & 29.13 & 2     & 100   & 153.25  & 29.76  & 5.5  & 100  & 159.65  & 30.96   & 10.5 & 96.5 \\
		DWA   & 185.1    & 29.85   & 15    & 95.5  & 187.6   & 34.15  & 53.5 & 83.5 & 195.60 & 30.07  & 83.5 & 77   \\
		MPC   & \textbf{134.45}   & \textbf{26.95}   & 0     & 100   & \textbf{138.16}  & \textbf{26.28} & 0    & 100  &\textbf{149.25}  & \textbf{29.33}  & 8.5  & 97.5 \\
		TEB   & 171.95   & 29.28  & 0     & 100   & 171.83  & 28.69 & 2.5  & 100  & 183.86  & 29.75   & \textbf{5.5}  &\textbf{ 99}   \\
		\cline{1-13} \hline
		\textbf{$v_{obs}$ 0.2m/s} \\
		\cline{1-13}
		CADRL & 150.46  & 29.70 & 11    & 97.5  & 184.03 & 36.51  & 44   & 90   & 181.14 & 34.71 & 53   & 86.5 \\
		\textbf{ARENA (Ours)} & 151.2    & 29.255  & 3     & 100   & \textbf{163.25} & 31.95  & \textbf{18 }  & \textbf{94}   & 161.60 & 31.45   & 30.5 & 92.5 \\
		DWA   & 202.36  & 30.16  & 48    & 85    & 189     & 34.68 & 65   & 80.5 & 205.25  & 31      & 120  & 58   \\
		MPC   &\textbf{ 147.97}   & \textbf{28.31}  & \textbf{2}     & 100   & 144.75 & \textbf{27.18} & 28.5 & 91   & 169.28 & 29.6    & \textbf{20}   & \textbf{94}  \\
		TEB   & 179.35   & 29.44  & 4.5   & 98    & 176.62 & 29.2   & 16   & 93 & 192.76 & 30.62   & 30   & 90.5 \\
		\cline{1-13} \hline
		\textbf{$v_{obs}$ 0.3m/s} \\
		\cline{1-13}
		CADRL & 153.3835 & 29.9625 & 43.5  & 86.5  & 185.96  & 37.168 & 73.5 & 75   & 206.195 & 40.015  & 130  & 62   \\
		\textbf{ARENA (Ours)}& 152.2    & 30.495  & \textbf{15.5}  & \textbf{94.5}  & \textbf{165.5}   & 32.1   & \textbf{29}   & \textbf{93}   & \textbf{170.275} & 33.875  & \textbf{65}   &\textbf{81}   \\
		DWA   & 215.62   & 31.195  & 55.5  & 82    & 188.5   & 34.65  & 75.5 & 76.5 & 204.975 & 31.985  & 120  & 56   \\
		MPC   & \textbf{149.05}   & 29.95   & 24.5  & 92    & 169.8   & \textbf{27.9}   & 40   & 88   & 178.745 & \textbf{30.475}  & 91.5 & 70.5 \\
		TEB   & 185.8    & \textbf{28.965}  & 23.5  & 91    & 212.54  & 33.43  & 44   & 85.5 & 221.215 & 31.65   & 96   & 68   \\
		\cline{1-13}
		\hline
		\hline
		& \multicolumn{4}{c}{\textbf{Overall Average}} & \multicolumn{4}{c}{\bfseries Empty Map} & \multicolumn{4}{c}{\bfseries Office Map} \\ 
		\cline{1-13}
		CADRL & 170.511   & 33.381   & 41.33 & 88.05 & 181.34  & 33.38  & 49.1 & 85.8 & 159.68  & 31.03   & 33.5 & 90.2 \\
		\textbf{ARENA (Ours)}& 158.57   & 30.99   & \textbf{19.88} & \textbf{94.61} & 166.96  & 30.9   & 23.7 & 92.7 & \textbf{150.18}  & \textbf{29.55}   & \textbf{16}   & \textbf{96.4} \\
		DWA   & 197.11   & 31.972  & 70.66 & 77.1  & 210.78  & 31.97  & 65.6 & 79.1 & 183.43  & 32.48   & 75.6 & 75.1 \\
		MPC   & \textbf{153.49}   & \textbf{28.443}  & 23.8  & 92.5  & \textbf{162.7}  & \textbf{28.4}   & 26.7 & 91.3 & 144.29  & 27.38   & 21   & 93.7\\
		TEB  & 188.43   & 30.115  & 24.88 & 91.83 & 202.75  & 30.11  & \textbf{18.6} & \textbf{94.1} & 174.12  & 29.57   & 31.1 & 89.5\\ \cline{1-13}\cline{1-13}\hline \hline
		
	\end{tabular}
	
	\label{quanttable}
	
\end{table*}

\subsection{Quantitative Evaluations}
Table \ref{quanttable} lists the quantitative evaluations of our experiments for all agents. In total, we conducted 100 test runs for each planner and each map. We evaluate the efficiency of path planning by calculating the average distance traveled, the average time to reach the goal and the collision rates with all objects. In addition, we define the success rate of all runs when the goal is reached with less than two collisions. We set the timeout to 3 minutes. To evaluate the capability of all planners to cope within highly dynamic environments, we test all planners on scenarios with 5, 10 and 20 dynamic obstacles, and three different obstacle velocities (0.1,0.3 and 0.5 m/s). Subsequently, the success rate, collision rate, path lengths and time to reach goal over the number of obstacles are plotted in Fig. \ref{quantgraph1} and \ref{quantgraph2}. 
\subsubsection{Safety and Robustness}
As expected, the learning-based planners CADRL as well as our DRL agent can cope with a higher number of dynamic obstacles and still manages to keep a success rate of nearly 94 \% with 10 obstacles whereas the well-performing MPC planner and the DWA planner only attain 91\% and 80.5\% respectively for $v_{obs}=0.2 m/s$. The gap is even more evident when increasing the obstacle velocities as can be observed in Fig. \ref{quantgraph1} and \ref{quantgraph2} and Table \ref{quanttable}. For the test runs with a dynamic obstacle velocity of 0.3 m/s, our planner outperforms all planners in terms of collision rate and success. In the most difficult scenario with 20 obstacles, our planner still manages a 81 \% success rate compared to 70.5 and 68 \% for MPC and DWA respectively while almost half of DWA's runs failed (56 \% success).
Out of the conventional approaches, TEB and MPC achieve best overall results with approximately 92 \% success rate.
MPC achieves the best results among those 3 conventional approaches with a success rate of over 88 and 70.5 \% for 10 and 20 dynamic obstacles of 0.3 m/s respectively. The superior performance by our DRL-based planners in highly dynamic environments in terms of success rate and safety is clearly observed. It attains a success rate of over 81 \% even with 20 obstacles, whereas these scenarios cause a rapid performance drop for the model-based approaches.
CADRL accomplishes similar performance in highly crowded environments but fail when obstacle velocities are increased. Whereas it attains 95 \% and 86.5 \% success in environments with $v_{obs}$ is 0.1 and 0.2 m/s respectively, success drops to 62 \% when $v_{obs}$ = 0.3 m/s.
This is due to the fact that CADRL was trained on a specific obstacle velocity of 0.2 m/s, which makes it inflexible for highly dynamic situations.
Whereas conventional planners cope well in scenarios with slow moving obstacles, their performance declines in environments with $v_{obs} < 0.2$. With increasing number and velocity of obstacles, the superiority of our ARENA planner in terms of success rate and safety becomes more evident. 
\subsubsection{Efficiency}
In terms of efficiency, MPC is the most efficient planner, taking less time and shorter trajectories compared to the other approaches. As stated in Table \ref{quanttable}, MPC require an average of 153.49 s with an average path length 28.44 meters, whereas CADRL and our ARENA planner require 170.5 and 158.57 s while the average path length is 33.38 and 30.99 meters respectively. 
However, similar to the success and collision rates, a decline in efficiency for all model-based planners can be noticed for scenarios with increased obstacle numbers and velocities. Our ARENA planner, outperforms all planners in terms of efficiency in scenarios with 10 and 20 obstacles and a $v_{os}$ of 0.2 and 0.3 m/s. 
\subsubsection{Overall Performance}
To compare our approach against all planners, we calculate the average performance for each metric (efficiency, robustness and safety) by dividing the respective overall values of the planners with the values from our DRL-based planner. The results are percentages indicating the relative performance to our ARENA planner. Subsequently, we added all percentages to attain the overall performance metric. The results over $v_{obs}$ are plotted in Fig. \ref{quantgraph1} (f) and \ref{quantgraph2} (f).
Whereas, for $v_{obs}=0.1 m/s$, MPC and TEB achieve a higher performance, our planner outperforms all planners for higher obstacle velocities. Notable is the high performance of the MPC planner for $v_{obs}$ at 0.1 m/s and 0.2 m/s in the office map but the rapid decline for $v_{obs}$ at 0.3. On the empty map this decline already occurs at $v_{obs}$ at 0.2 m/s. This is due to the fact that model-based planners perform worse on empty map due to their high dependence on the global planner and a global map with static obstacle as marker points. CADRL achieves similar performance to our ARENA planner. However, one main limitation of CADRL is that it requires the exact positions of obstacles and can only cope with circular obstacles. In contrast, our planner was trained solely on laser scan observations and thus can accommodate any kind of obstacles without prior knowledge of its exact position. This makes our planner more flexible for real world application.

%% file: ieeeconf/content/5-conclusion.tex
\section{Conclusion}
In this paper, we proposed a framework to integrate DRL-based local planners into long range navigation. Our platform can be used to directly train and deploy DRL-based algorithms as local planners combined with classical global planners. Our proposed intermediate planner will generate way-points and has shown success in being more efficient at replanning than the existing ROS navigation stack and providing reasonable way-points when the robot is stuck.
We integrate various conventional model-based as well as learning-based obstacle avoidance algorithms into our system and propose a DRL-based local planner using a memory-enhanced A3C. Subsequently, we evaluate all planners in terms of safety, efficiency and robustness. Results demonstrate the superiority of learning-based planners compared to conventional model-based ones on all metrics. Future work includes the integration of semantics into the DRL for behavior adaption and more sophisticated algorithms for our intermediate planner. Therefore, we aspire to extend our approach by including the intermediate planner into the reinforcement learning training process.

% \begin{table}[H]
% 	\setlength{\tabcolsep}{2pt}
% 	\renewcommand{\arraystretch}{0.7}
% 	\centering
% 		\caption{Rewards and Penalties}
% 	\begin{tabular}{lllll}
% 		\hline
% 		Event  & Value   & Description    \\ \hline
% 		Goal    & 100        & Agent reaches goal          \\ 
% 		Towards goal    &0.1       &Agent moves towards goal        \\ 
% 		Col-Human &-100        & Agent collides with human          \\ 
% 		Col-Robot &-80        & Agent collides with robot          \\ 
% 		Away from goal    & -0.2       & Agent moves away from goal        \\ 
% 	    Hit Static Object    & -10      & Agent hits static obstacle        \\ 
% 		\hline
% 	\end{tabular}

% 	\label{actions}
% \end{table}
\section{Appendix}

\begin{table}[!h]
\centering
	\setlength{\tabcolsep}{1.5pt}
	\renewcommand{\arraystretch}{1.05}
		\caption{Hyperparameters for Training}
	\begin{tabular}{ccp{4.1cm}}
		\hline
		Hyperparameter  &Value& Explanation    \\ \hline
		Mean Success Bound & 1        & Training considered done if mean success rate reaches this value           \\ 
		Discount Factor & 0.9    & Discount factor for reward estimation (often denoted by gamma)   \\ 
		Learning Rate   & 0.00025        &Learning rate for optimizer           \\ 
		Epsilon Max Steps   & $10^5$        &Steps until epsilon reaches minimum         \\ 
		Epsilon End   & 0.05        &Minimum epsilon value    \\ 
		Batch Size   & 64        &Batch size for training after every step          \\ 
		Memory Size  & $10^64$        &Last X states will be stored in a buffer (memory), from which the batches are sampled        \\ 
		Trade-off-factor $\lambda$  & 0.95        &Trade-off-factor between bias and variance      \\ 
		Clip-range & 0.2        & Clipping range for the value function to mitigate gradient exploding     \\ 
		Maximum Gradient   & 0.5        &Maximum value to clip the gradient     \\ 
		\hline
	\end{tabular}

	\label{tablehyper}
	
\end{table}